# Adaptive Learning of Region-based pLSA Model for Total Scene Annotation

*Yuzhu Zhou\**
Department of Computer Science and Engineering
The Chinese University of Hong Kong
zhouyuzhu00@gmail.com

*Le Li\**
International School
Beijing Univ. of Posts & Telecoms
buptlile@gmail.com
(\* indicates same contribution)

*Honggang Zhang*
Pattern Recognition and Intelligent System Laboratory
Beijing Univ. of Posts & Telecoms
zhhg@bupt.edu.cn

*Abstract*—in this paper, we present a region-based pLSA model to accomplish the task of total scene annotation. To be more specifically, we not only properly generate a list of tags for each image, but also localizing each region with its corresponding tag. We integrate advantages of different existing region-based works: employ efficient and powerful JSEG algorithm for segmentation so that each region can easily express meaningful object information; the introduction of pLSA model can help better capturing semantic information behind the low-level features. Moreover, we also propose an adaptive padding mechanism to automatically choose the optimal padding strategy for each region, which directly increases the overall system performance. Finally we conduct 3 experiments to verify our ideas on Corel database and demonstrate the effectiveness and accuracy of our system.

*Keywords-Annotation; pLSA; region-based; CEDD*

## I. INTRODUCTION

New technology application often inspires the researchers to come out new solution to the traditional problems. The success of Flicker gives researcher a new way to explore how to efficiently and accurately search images with the help of tags information. And the integration of low-level visual feature and tags information can considerably reduce the semantic gap in traditional Content-based image retrieval (CBIR) system. However, how to generate proper tags to describe the contents in each image still remains a challenge problem. Generally speaking, there are three categories of approaches for image annotation. The first solution is to single labeling for each image, which means using one tag to describe the dominant information at one image [1]. This approach, to some extent, can be treated as classification problems. The second approach is to assign a list of tags without localizing their corresponding spatial positions in the original image, e.g. [5]. The last approach provides both a few tags as well as each tag's spatial location in image [2]. Basically, the essence of third approach is to build a correspondence between each region's low-level feature and semantics (i.e. tag) and it helps us better understanding the scene of each image.

Our work is focused on the third approach, which inspires us to concentrate on the study on regions. [3] proposes a Region-based retrieval system that first implemented a strong segmentation method named JSEG, and then extract features from these separated regions. However, their adoption of decision-tree for classification has poor generalizability performance and needs complicated configuration for each specific database. On the other hand, in [4], a pre-segmentation method is used to extract patches from images. In order to ensure each region contain single topic, they use over-segmentation and that really reduce the integrated information of the target object.

In this system, we propose a region-based pLSA system for full-scene annotation of each image. It consists of four major blocks: segmentation, adaptive padding, low-level feature expression and annotation. In segmentation block, similar with [3], we use a state-of-art method called JSEG to segment each image into different regions, which takes advantage of color and texture information for segmentation so that we can achieve a good segmentation effect but in a comparatively efficient way. In the $2^{nd}$ block, an adaptive padding mechanism will be used to decide whether each region should be represented as pad-origin or pad-zero strategy so that we can achieve a better feature expression and annotation performance. In the third part, our refined version of Color and Edge Directivity Descriptor (CEDD) [10], a useful and efficient multiple feature descriptor, will be adopted to represent each padded region's low-level feature. At the last block, pLSA algorithm will be used to generate each region's tags information according to their low-level features in latent spaces. And with the help of these tags, we can achieve a full-scene annotation in image. The contributions of our work are the following two aspects: (1) adaptive padding mechanism can find each region's best padding strategy after segmentation, which directly increases annotation accuracy and improve the generalizability of system; (2) we build an efficient and powerful region-based framework with efficient segmentation performance and statistical model.

The rest of paper is organized as follows: section 2 presents the background knowledge on JSEG, CEDD. Section 3 describes the detail functionality of each component in our system, such as region-based pLSA model, adaptive padding and annotation mechanism. In section 4, three experiments are conducted. And their results will be presented and analyzed. Finally, we conclude our work and discuss the future work in section 5.

## II. BACKGROUND

### A. Image Segmentation

The general purpose of segmentation is to divide the original image into different regions where each region shares the same local patter. So, ideally, one region should correspond to one object. However, due to the limitation of low-level-feature-based segmentation techniques, we cannot achieve such kind of semantic correspondence. So, over-segment is often adopted since at least each region represents the partial information of one object.

JSEG, as a state-of-art unsupervised segmentation method, consists of two steps. Firstly, color information of each image is quantized to form a class map, which maps the original pixels from their corresponding color class labels. Secondly, 'J Value' is used in spatial segmenting to estimate complexity of each region with the class-map. So, basically, 'J Value' can be treated as the criterion for good performance.

Meanwhile, the parameter, threshold of merge (TM), in the color quantization step considerably affects the performance of JSEG. TM value can directly decide the degree of segmentation on original image. Empirically, we adopt TM=0.55 in this paper.

### B. CEDD for region representation

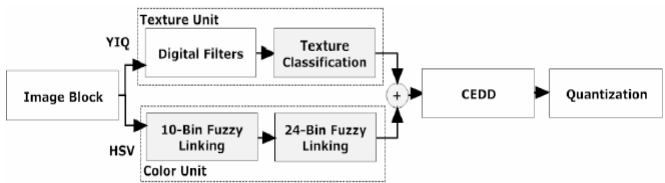

Figure 1.  Block diagram of CEDD

Color and Edge Directivity Descriptor (CEDD) is a newly-proposed low-level multi-feature descriptor which properly integrates both color and texture and forms a 144-demension feature vector. CEDD has two significant advantages: 1) computation efficient compared with most MPEG-7 descriptors; 2) make a reasonable combination of color and texture features rather than simply concatenate them together. The block diagram of CEDD can be seen in figure 1.

Traditional CEDD algorithm also provides a quantization functionality to reduce the 144 byte vector into 432 bits. But in our experiment we abandon this quantization part since we want to precisely represent the feature information in each descriptor element using 8 bits. And original 3 bits for each bin is too low.

### III. SYSTEM DESCRIPTION

#### A. General Introduction

We propose a region-based pLSA model with adaptive learning mechanism, which aims to segment scene images into several regions and label each region with related topics. It contains three stages: pLSA model training, adaptive padding and total scene annotation stage (i.e. test stage). Firstly, once each input images are segmented into dozens of regions, the system generates several pLSA training models in different padding strategy situations. Then, at adaptive padding stage, we make useful of the output at first stage to build the relationship between each region's feature expression and its corresponding optimal padding method. The output of adaptive padding stage is the SVM training model describing the feature-padding relationship. Finally, we can take advantage of these outputs from previous two stages to achieve total scene annotation of test image.

### B. pLSA training model

Probabilistic latent semantic analysis (pLSA) [7] is a statistical model based on the assumption of existence of a hidden variable (latent aspect) $z_k$ in the generative process of each term $f_j$ in specific region's feature descriptor $r_i$. Basically, a term may be derived from many regions, which indicates that region may contain multiple aspects. Given hundreds of labeled training regions, our proposed region-based pLSA algorithm is mainly designed to build the training model to help extract the semantics, i.e. predict the most probable class of each region. And the graphical representation of pLSA model can be seen in figure 2.

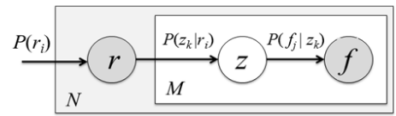

Figure 2.  the graphic representation of pLSA in our system

$P(z_k|r_i)$ and $P(f_j|z_k)$ are two conditional distribution of the pLSA's model parameters. They are estimated by an EM algorithm on a set of training regions. $P(f_j|z_k)$ characterizes each aspect, and remain valid for regions out of the training set. $P(z_k|r_i)$, on the contrary, is only relative to the training regions and doesn't contain any prior information to an unseen region.

Initially, each region's feature will be represented as a 144-dimension vector using CEDD. So j-th element of this feature vector will be observed as word $f_j$. And each $f_j$ will correspond to a hidden variable $z_k$ in statistical latent model, which, essentially, represents the topic information of each element. For example, if all the regions belong to 8 classes, the number of the hidden variable z is 8, i.e. K=8.

In order to get the probability mapping between R (region) and Z (hidden variable), we need to solve following joint probability,

$$P(r_i, f_j) = P(r_i) \sum_{k=1}^{K} P(z_k|r_i) P(f_j|z_k).$$

The parameters here are $P(z_k|r_i)$ and $P(f_j|z_k)$. The calculation of the model parameters is through EM algorithm, which maximizes the log-likelihood of the observed data through two steps: 1) E-step: compute the conditional distribution of $P(z_k|r_i, f_j)$ from the previous estimation; 2) M-step: update the value of $P(f_j|z_k)$ and $P(z_k|r_i)$ according to the new $P(z_k|r_i, f_j)$ data. So the essence of EM algorithm is to find the optimal value of unseen data through fixed known parameters.

$$\mathcal{L} = \sum_{i=1}^{N} \sum_{j=1}^{M} n(r_i, f_j) \log P(r_i, f_j).$$

Where $n(r_i, f_j)$ is the count of feature element $f_j$ in region $r_i$.

At two steps at EM algorithm, we need to solve:

1. E-step.
$$P(z_k|r_i, f_j) = \frac{P(z_k|r_i)P(f_j|z_k)}{\sum_{k=1}^{K} P(z_k|r_i)P(f_j|z_k)}.$$
2. M-step.
$$P(f_j|z_k) = \frac{\sum_{i=1}^{N} n(r_i, f_j)P(z_k|r_i, f_j)}{\sum_{i=1}^{N} \sum_{j=1}^{M} n(r_i, f_j)P(z_k|r_i, f_j)}.$$
$$P(z_k|r_i) = \frac{\sum_{j=1}^{M} n(r_i, f_j)P(z_k|r_i, f_j)}{\sum_{j=1}^{M} n(r_i, f_j)}.$$

### C. Adaptive Padding System

According to the research of [8], a general problem of Region-based models is how to extract features from arbitrary-shaped regions. The problem is noticeable since it directly determines the region representation, which considerably affects the following system performance. We classify current padding strategies into three categories: (1) Pad unvalued part with zero (i.e. black)[8]. This strategy is very simple, but it may introduce many unwanted information into feature expression. In this paper, we donate this strategy as pad-Z; (2) Pad with mirror. [9] uses a technique named POCS-ER to deal with the problem. However, POCS-ER is not practical for many situations since it cannot suit for regions that contain lots of unvalued parts. (3) Pad with original information. [6] proposes an easy and effective way to solve the problem that padding with original information from the original image. We donate this strategy as pad-O. From [6], we can see that pad-O outperforms pad-Z in many situations.

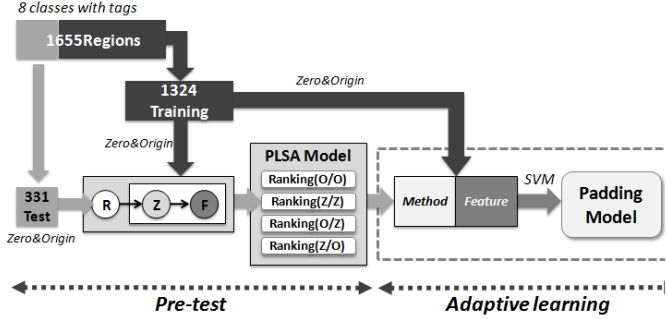
Figure 3.  Block diagram of adaptive padding.

However, every padding strategy has its limitation and we cannot conclude that one strategy is surely better than the rest. In order to take full advantage of various methods, in our system we propose an adaptive padding system that can choose the best padding strategy for both training and test region with options: pad-Z or pad-O. Its basic idea is to "pre-test" same region with different padding strategy using the pLSA training model we build in section 3.2. The topics of test samples are actually known (i.e. labeled). Then we calculate the accuracy of each padding strategy on each class and find its corresponding strategy. As a consequence, we can adopt this correspondence relationship to automatically learn training model in SVM. Here, we divide the adaptive padding part into two processes as shown in figure 3: 1) pre-test; 2) adaptive learning.

Given 8 classes of regions, the purpose of pre-test is to find the best padding strategy for each class. In our system, we have 1655 labeled regions. Among these 1655, we use 1324 regions for training and the rest 331 for pre-test. And each region has two versions of expression: pad-Z and pad-O. So, we can build two pLSA training models from these 1324 samples: pLSA-Z model and pLSA-O model. Then, we will input two versions of 331 samples into these two pLSA models, respectively. And calculate the accuracy of each class under different padding strategies and find its optimal strategy. Basically, each class of regions will get four probability rankings tables after pLSA estimation since both training and test samples have 2 padding options. And each probability ranking table indicates the ordered likelihoods of potential classes in this padding strategy. For example, as to Flight class, it can receive 4 probability rankings tables under the circumstances of: 1) training model is pad-O, test sample is pad-O(donate O/O); 2) training model is pad-Z, test sample is pad-Z(donate Z/Z); 3) training model is pad-O, test sample is pad-Z(donate O/Z);4)training model is pad-Z, test sample is pad-O(donate Z/O). Assuming in O/O case, the top likelihood output of one pre-test region in probability ranking table is Flight. Then it indicates a correct prediction, otherwise wrong.

After pre-test, each class will find its optimal padding strategy, which means that we can build a mapping relationship between each class's feature expression and its corresponding padding strategy. According to this correspondence, the system can automatically label the class number of each feature vector during SVM training stage without human participation. That's why our system is called adaptive learning. By doing so, we can implicitly change SVM from supervised into unsupervised method with high generalizability. The output of adaptive learning is the padding model that will be used in total scene annotation stage.

### D. Total scene annotation

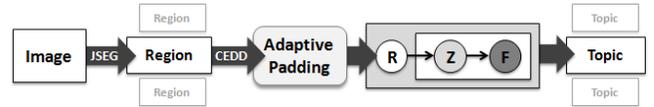
Figure 4.  Block diagram of Total scene annotation

Given a test image, our proposed system provides simultaneously segmentation, annotation and localizes the corresponding tags to the certain area of the image (shown in figure 4). Based on the two types of training models (pLSA and SVM) derived from section 3.2 and 3.3, we can make a total scene annotation to input test image. Firstly, the test image is segmented into various regions. The threshold of JSEG algorithm is 0.55 which shows best performance among several experiments as indicated in [6]. Secondly, low-level feature vectors are extracted from each region after adaptive padding. Under the representation by CEDD descriptor, a number of 144-dimensions vectors are inputted into the specific region-based pLSA model. For example,

Each region corresponds to one probability ranking list, and comparing with a threshold, only one tag is associated to the region. With these labeled regions, the semantics can be extracted from the image and localized in the certain area.

Please punctuate a displayed equation in the same way as ordinary text but with a small space before the end punctuation.

## IV. EXPERIMENTAL RESULTS

### A. Build PLSA model

In this experiment we will find each class's optimal padding method through pre-test. We test our region-based pLSA model on 5 scene categories from Corel database: butterfly, flight, flower, mountains and cats; totally shared with 8 classes of objects: butterfly, leaves, flower, flight, sky, mountains, plants and cats. There are 1344 images and we use a segmentation threshold 0.55. Details can be seen in Table.1). Then, we manually choose 1655 'proper' regions with comparatively large size and containing meaningful object information from whole segmented regions. Among them, 1324 regions are used as training samples and the rest 331 for pre-test.

TABLE I. SAMPLES INFORMATION OF OUR EXPERIMENT

| Scene | Images | Regions | Reg/Img |
|---|---|---|---|
| Butterfly | 104 | 1882 | 18.1 |
| Flight | 207 | 1694 | 8.2 |
| Flower | 374 | 5264 | 14.1 |
| Mountain | 243 | 2892 | 11.9 |
| Cats | 416 | 5003 | 12.0 |
| **TOTAL** | 1344 | 16735 | |

Our system provides two padding method: pad-O and pad-Z. Hence for both the training set and test set, there both have two versions of expression. Totally we have four probability distribution results (i.e. the probability ranking table). The CEDD descriptor represents each region as a 144-demensions vector. For the 1324 labeled regions, we have a 144x1324 matrix for $P(f_j|r_i)$ and 8x1324 matrix for $P(z_k|r_i)$. Equally for the 331 test regions, we have a 144x331 matrix with no topic information for test.

Table 2 shows the number of right-classifications through different padding methods for training and test. The bold-labeled number in each class indicates the best padding strategy for this class. For example, as the Butterfly class, among 25 test samples, training sample with pad-O and testing sample with pad-Z, donated as O/Z, receives the best accuracy with 12 right-classifications (i.e. 48%). Since O/Z and O/O occupy almost all optimized padding strategies, we only list the details classification accuracy in percentage of these two, as indicated in Table3 (the results of Z/O and Z/Z are not provided here). So, in our work, we can simply adopt two pLSA training model under O/O and O/Z circumstance. And those regions, belonging to the following 4 classes: butterfly, leaves, flower and flight, will correspond to O/Z strategy while the rest regions are O/O.

### B. Equations

In this experiment, we will use data to demonstrate the superiority of our adaptive padding mechanism. From the result in previous experiment, we know that the ideal padding strategy is that the first 4 classes' regions should use O/Z and the rest O/O, as indicated in Table 4 (a). However, due to the classification error, it is very hard to achieve such performance. But, from Table 4(b), we can see that our method still has very good classification performance and 199 test samples are rightly classified into desired padding strategies among 331 samples, which approximate the ideal situation 203/331.

TABLE II. RESULT OF PRE-TEST VIA PLSA MODEL

| Classes | Train | Test | O/O | Z/Z | O/Z | Z/O |
|---|---|---|---|---|---|---|
| 1.Butterfly | 101 | 25 | 9 | 7 | 12 | 7 |
| 2.Leaves | 192 | 48 | 44 | 33 | 48 | 34 |
| 3.Flower | 173 | 43 | 13 | 12 | 17 | 11 |
| 4.Flight | 106 | 27 | 2 | 4 | 6 | 0 |
| 5.Sky | 274 | 69 | 54 | 52 | 9 | 54 |
| 6.Mountain | 193 | 48 | 16 | 9 | 8 | 16 |
| 7.Plants | 126 | 31 | 18 | 14 | 1 | 20 |
| 8.Cats | 159 | 40 | 32 | 30 | 5 | 32 |
| **TOTAL** | 1324 | 331 | 188 | 161 | 106 | 174 |

TABLE III. PRE-TEST DETAILS ON O/O AND O/Z SITUATIONS

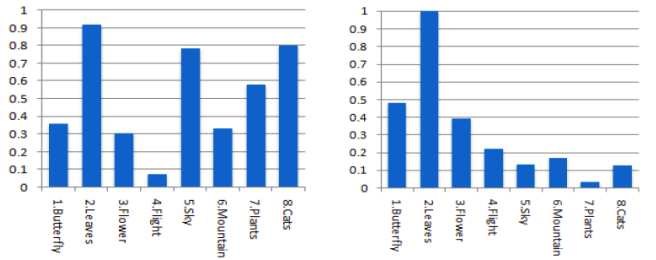

Moreover, compared with solely O/O or O/Z strategy for all 331 samples in section 4.1, our adaptive increase the performance by 7% and 87% in terms of total right classified number, respectively. The reader may notice the improvement for O/O is very small (7%). But it doesn't reduce the importance of our adaptive padding strategy. The reason is explained as follows: even through O/O performs well in almost all classes, what we concern most is how to find the optimal padding strategy for each class. As to the first four classes, O/Z outperforms O/O greatly. So, we surely incline to assign O/Z strategy for region belonging to first four classes.

### B. Some Common Mistakes

In this experiment, we will verify merit of pLSA model for semantic information extraction after padding method is determined for each test region. Here we use 60 images as test samples. And the average number of region per image is 8. Every scene has 12 images. We evaluate the annotation

TABLE IV. OUR PADDING METHOD

(a) Ideal Padding Strategy

| | 1 | 2 | 3 | 4 | 5 | 6 | 7 | 8 |
|---|---|---|---|---|---|---|---|---|
| 1. | 48 | 16 | 20 | 8 | 0 | 8 | 0 | 0 |
| 2. | 0 | 100 | 0 | 0 | 0 | 0 | 0 | 0 |
| 3. | 35 | 12 | 40 | 5 | 0 | 9 | 0 | 0 |
| 4. | 19 | 19 | 0 | 22 | 0 | 33 | 4 | 4 |
| 5. | 1 | 4 | 6 | 1 | 78 | 1 | 1 | 6 |
| 6. | 6 | 0 | 4 | 10 | 0 | 33 | 6 | 40 |
| 7. | 13 | 6 | 0 | 13 | 0 | 3 | 58 | 6 |
| 8. | 20 | 0 | 0 | 0 | 0 | 0 | 0 | 80 |

Total: 203/331

(b) Our Adaptive Padding Method

| | 1 | 2 | 3 | 4 | 5 | 6 | 7 | 8 |
|---|---|---|---|---|---|---|---|---|
| 1. | 48 | 19 | 17 | 9 | 0 | 7 | 0 | 0 |
| 2. | 0 | 96 | 0 | 0 | 0 | 0 | 4 | 0 |
| 3. | 34 | 11 | 40 | 6 | 0 | 9 | 0 | 0 |
| 4. | 21 | 17 | 0 | 22 | 0 | 32 | 5 | 3 |
| 5. | 3 | 7 | 7 | 1 | 77 | 1 | 0 | 4 |
| 6. | 13 | 5 | 6 | 10 | 0 | 31 | 6 | 29 |
| 7. | 12 | 6 | 0 | 13 | 0 | 3 | 58 | 7 |
| 8. | 20 | 0 | 0 | 0 | 0 | 0 | 0 | 80 |

Total: 199/331

performance by comparing the recommended tags given our system with the original manual annotations in Corel. Similar with [2], we use recall-precision and F-measures as metric. Table 5 lists the details annotation result for 8 classes. As we can see from this table, our model achieves very high annotation result. That is mainly attributed to the fact pLSA model basically is more suitable for semantic information extraction by introducing the hidden variables.

TABLE V. RECALL-PRECISION VALUES FOR ANNOTATION

| Classes | Prec | Rec | F |
|---|---|---|---|
| 1.Butterfly | 0.97 | 0.72 | 0.83 |
| 2.Leaves | 0.86 | 0.85 | 0.85 |
| 3.Flower | 0.56 | 0.75 | 0.64 |
| 4.Flight | 0.89 | 0.86 | 0.87 |
| 5.Sky | 0.88 | 0.93 | 0.90 |
| 6.Mountain | 0.67 | 0.69 | 0.68 |
| 7.Plants | 0.87 | 0.83 | 0.85 |
| 8.Cats | 0.92 | 0.97 | 0.94 |
| **Mean** | **0.83** | **0.83** | **0.82** |

Figure 5 shows several examples of annotation obtained by our proposed model. Each region is assigned with no more than one tag. As we can see, those classes of objects trained in our system have been right presented.

Figure 6 demonstrates several examples of full scene annotation results. Those near-by regions belonging to same class will be annotated with same color for simplicity purpose. The red tags in the each image are the wrong annotation results. Thanks to the good segmentation algorithm, adaptive padding strategy and semantics extraction method, our annotation can reach a good result.

V. CONCLUSION

In this paper, we propose a region-based pLSA framework with adaptive learning to accomplish the task of total scene annotation. The JSEG part can guarantee the accurate representation of each region in an efficient way. pLSA model can help capturing semantic information for each region. Meanwhile, the proposed adaptive learning mechanism considerably improves the feature expression and later annotation accuracy. From the experiment, we can achieve a good correspondence result between each region with its tag and localize this relationship in original image.

In the future, we will take the tags information from users as another input to our system. It requires us make a good integration of existing tags and low-level feature in region-based model and find good methods to deal with noisy tags.

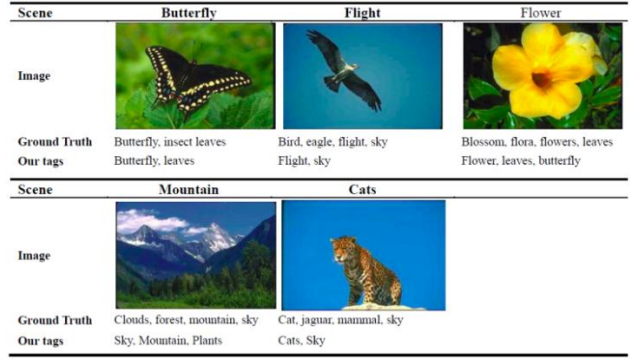

Figure 5. Comparison of our annotations with ground-truth tags

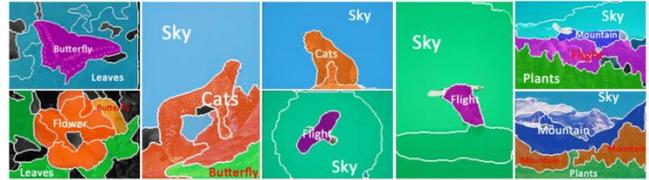

Figure 6. Examples of total scene annotation using our model

ACKNOWLEDGMENT

This paper was partially sponsored by the grants from the Fundamental Research Funds for the Central Universities, 111 Project of China (B08004) and Scientific Research Foundation for the Returned Overseas Chinese Scholars, State Education Ministry.